\documentclass[letterpaper, 10 pt, conference]{ieeeconf}  

\IEEEoverridecommandlockouts                              

\usepackage[caption=false, font=footnotesize]{subfig}
\usepackage{graphicx}
\usepackage{graphics} 
\usepackage{amsmath} 
\usepackage{amssymb}  
\usepackage{courier}
\usepackage{url}

\title{\LARGE \bf
HALO: Heterogeneous Allocation via Localized Observations for the Vehicle Routing Problem
}

\author{Andrew Meighan$^{1}$, Hyungsub Kim$^{2}$, and Or D. Dantsker $^{3}$
\thanks{Corresponding author: Andrew Meighan ({\tt\small ameighan@iu.edu})}
\thanks{$^{1,3}$Authors are with the Department of Intelligent Systems Engineering,
        Indiana University Bloomington}%
\thanks{$^{2}$Author is with the Department of Computer Science,
        Indiana University Bloomington}%
}

\begin{document}

\maketitle

\thispagestyle{empty}
\pagestyle{empty}

\begin{abstract}

Scalable robotic fleets have become increasingly popular for various applications such as package delivery, warehouse management, and military operations. 
Prior fleet control algorithms solve centralized routing problems with up to $1{,}000$ tasks in controlled environments, yet they fail to consider realistic constraints such as limited observation and communication ranges typical of decentralized fleets.
Thus, deploying existing fleet control algorithms into real-world settings is currently infeasible.

To tackle this, we propose Heterogeneous Allocation via Localized Observations (HALO) to solve the Vehicle Routing Problem (VRP). 
HALO is a hybrid method that splits the VRP into allocation and routing portions to provide onboard, real-time solutions to robots in dynamic environments. 
During the allocation phase, HALO utilizes a heterogeneous graph neural network framework with unique message passing layers to explicitly separate the learning of spatial distributions and task-to-robot compatibility. 
HALO then leverages heuristic methods to solve the smaller-scale, single-robot routing problems. 

Evaluation results on a partially observable, online variant of the VRP show HALO significantly outperforms the heuristic baseline while maintaining similar solution quality to an all-knowing offline variant of HALO. 
As the proportion of hidden tasks increases, the decentralized fleet maintains average makespans within 4.8\% of a centralized fleet with no environmental constraints.
Notably, as operations scale to $100$-robot fleets, the decentralized architecture actually surpasses this offline variant. 
While HALO is explicitly designed for partially observable environments, it imposes no strict upper bound on the observation space allowing us to test HALO on the traditional static, single-depot VRP. 
Here, HALO outperforms state-of-the-art architectures strictly optimized for the static variant of the VRP by up to 14.06\% on smaller-scale problems while remaining competitive at larger scales up to $1000$ tasks. 
Throughout all testing, this framework maintains the quickest execution times which emphasizes its potential for large-scale, real-time deployment. 

\end{abstract}


\section{Introduction}
The deployment of robotic fleets has become increasingly popular when solving large-scale coordination and control problems like last-mile logistics and warehouse fulfillment. 
These missions consist of huge numbers of sub-tasks; thus, fleet size directly affects potential solution quality.
For instance, Amazon deploys over one-million ground robots in warehouses to manage the huge volume of package orders that arrive every day \cite{dresser25amazondeployment}. 
While this type of automated fulfillment occurs in fairly structured conditions, large-scale coordination is also required in more dynamic environments. 
The Defense Advanced Research Projects Agency (DARPA) OFFSET program focused on developing decentralized fleets of more than two-hundred small uncrewed aerial vehicles for offensive swarm-enabled tactics \cite{chung23offset}.

Coordinating decentralized fleets requires an algorithm that minimizes the total travel cost while generating routes that adapt to changing environmental conditions (e.g., neighboring robots moving dynamically or dynamic moving obstacles).
This coordination challenge is commonly referred to as the Vehicle Routing Problem (VRP) which envelops many unique routing variants.
When modeled as a Network-Distributed Partially Observable Markov Decision Process (ND-POMDP) \cite{nair2005ndpomdp}, fleets lose the guarantee of global state knowledge.
Instead, robots must coordinate local observations with neighboring robots to build a shared understanding of the environment and cooperatively solve tasks (e.g., deliver packages on crowded public roads). 

Algorithms must also satisfy real-time operational constraints such as update speed, and limited communication and sensing ranges. 
In fact, for unmanned aerial vehicles (UAVs) in large fleets, ad-hoc communication networks are typically limited to distances between hundreds of meters and a few kilometers \cite{hayat16networksurvey}. 
Application and sensor suite heavily impact a robot's observation range. 
Precise information from visual cameras degrades past 20 meters \cite{keselman17camera}, while acoustic sensors and thermal cameras may provide long-range detection of potential targets up to 2 kilometers~\cite{zipline_daa} and 8 kilometers \cite{allison16wildfiredetection}, respectively.

Since the VRP often considers additional operational constraints like vehicle capacity and dynamic task sets, many studies initially address the Multiple Traveling Salesman Problem (mTSP) which focuses solely on the routing problem. 
As these problems scale, traditional optimization techniques \cite{baldacci2008exactalgoforvrp, gurobi, pecin2017bcpforcvrp, woulda24pyvrp} fail to reach solutions in time frames required for real-world applications. 
As noted in \cite{hu2020rlhybrid}, state-of-the-art exact solvers like Gurobi \cite{gurobi} take more than one hour to converge for routing problems with 10 robots and 100 tasks which is far too slow for real-world application. 
Heuristic methods such as \cite{shaw98lns, glover1989tabu1, baker2003gaforvrp, dorigo1997acofortsp} relax constraints on optimality, but face similar problems at scale because their iterative approach cannot investigate enough of the problem space.

In search of methods that provide quality solutions for real-time use, studies using machine learning methods to solve the mTSP and VRP have increased in prominence since~\cite{vinyals2017pointernetworks}. 
While results have shown promising results at solving static routing problems, each of these methods make a number of unrealistic assumptions regarding operational environments. 
The most common assumption, global observability, provides vehicles with access to an error-free observation of the entire state (e.g., positions of all robots and tasks in the environment)~\cite{hu2020rlhybrid, park2021schedulenet, cao2022dan, guo2024imtsp}.
Yet, in realistic operational conditions without centralized command, neighboring robots fall in and out of communication range and tasks enter and leave the observation radius resulting in variable observation sizes. 
Such dynamic operations cause problems for deep learning architectures constrained by fixed input and output dimensions.

To tackle these issues, a vehicle would need to store a unique model for each permutation of fleet size and task count.
Yet, constructing diverse sets of machine learning models is both burdensome and impractical, as fleet sizes and task types vary according to the applications.

To tackle the limitations of prior algorithms that solve the VRP, we propose Heterogeneous Allocation via Localized Observations (HALO).
HALO is a hybrid routing framework that enables onboard rerouting decisions for decentralized robots by splitting the VRP into allocation and routing stages. 
This framework avoids the high costs of long distance data transmission and allows robots to operate in remote environments without expensive infrastructure like cell towers.
Critically, HALO supports partial observability and variable fleet sizes by utilizing a heterogeneous graph neural network (GNN). 
Comprised of distinct message passing layers \cite{xu2019graphisomorphismnetworks, velickovic2018graphattentionnetworks}, HALO separates the learning of spatial node distributions from task-to-robot compatibility before merging the two representations into a single probability matrix for assignment. 

After task allocation, robots use a heuristic-based solver \cite{ortoolsrouting} to efficiently find near-optimal solutions to their individual subproblems.
The main contribution of this paper lies in the decentralized architecture rather than proposing a new optimization algorithm; thus, we adopt REINFORCE \cite{williams92reinforce}, a widely used policy gradient method used in the routing literature \cite{hu2020rlhybrid, cao2022dan, park2021schedulenet}, to train the network by minimizing the longest path in the fleet. 
Finally, to overcome the noisy reward signals inherent to training multi-robot fleets from scratch, we implement a curriculum learning strategy that guides learning by slowly increasing the variance of depot locations. 

HALO allows robots to construct localized subgraphs based on their limited communication and observation ranges. 
This enables robots to make team-oriented decisions without a centralized decision maker artificially eliminating assignment conflicts. 
HALO also generalizes across static single-depot and dynamic multi-depot environments without requiring retraining for different fleet sizes and task counts.

We evaluate HALO on two main VRP variants. 
First, we compare HALO with heuristic and learning-based methods on offline, single-depot VRP instances.
Results show a single HALO network trained on a fixed fleet size and task count outperforms state-of-the-art learning-based methods on smaller problem sizes while remaining competitive at larger scales even though the baseline methods strictly optimize for the static case. 
As the problem size increases, HALO maintains the quickest execution times.

We also validate the proposed method's ability to operate in real-time within dynamic, partially observable environments. 
In this case, robots are strictly limited by local observation and communication ranges and do not require a shared depot. 
Instead, the robots move around the environment, recalculating their discrete allocation at each step.
The results support the hypothesis that HALO can achieve near-optimal performance using only high-fidelity local information and inferred global context.

In summary, we make the following contributions:

\begin{itemize}

\item
\textbf{Decentralized Execution via Local Subgraphs.} 
To consider real-world constraints (e.g., limited communication and sensing ranges), we propose a scalable, decentralized execution strategy designed for partially observable environments. 
By allowing robots to construct localized subgraphs, we enable them to make fast, independent, team-oriented routing decisions based on local observations received from their neighbors.

\item 
\textbf{Heterogeneous GNN for Dynamic Environments.}
We construct a heterogeneous Graph Neural Network with distinct message passing layers to separate the learning of spatial node distributions from the task allocation. 
This unique architecture allows the framework to seamlessly adapt between static, single-depot benchmarks and dynamic, online environments without requiring network retraining for varying fleet sizes or task counts.

\item 
\textbf{High-Performance Decentralized Allocation.}
Extensive evaluations demonstrate that HALO significantly outperforms state-of-the-art baselines. 
In dynamic, partially observable environments, our decentralized method maintains makespans within a 5\% range regardless of hidden target percentage, and even surpasses a globally aware model for large fleet sizes (50, 100 robots) while consistently achieving sub-second replanning times suitable for real-world deployment. 
Furthermore, on traditional static benchmarks, HALO outperforms baselines strictly optimized for static cases by up to 14.06\%.

\end{itemize}

\section{Related Work}

\noindent\textbf{Traditional Optimization Techniques.}
Exact methods \cite{baldacci2008exactalgoforvrp, gurobi, pecin2017bcpforcvrp} systematically explore the VRP's solution space by creating a massive tree of potential route combinations. 
These algorithms use set partitioning and limited-memory cuts to significantly tighten the lower bounds and mathematically prune branches that cannot contain optimal solutions. 
Heuristic-based methods reduce computational complexity by giving up guarantees of optimality. 
Search methods \cite{shaw98lns, glover1989tabu1, glover1990tabu2} leverage tree-based search that iterates over an initial solution for a given time period. 
Other heuristic methods that have seen extensive use in routing problems include Genetic Algorithms \cite{baker2003gaforvrp, prins2004evolutionaryvrp, vidal2012geneticvrp} and Ant Colony Optimization \cite{dorigo1997acofortsp, yu2009improvedacoforvrp}. 
Yet, all of the traditional optimization-based algorithms share the following limitations: 
($1$) lack of scaling due to the exponential growth of the solution space (heuristic methods merely delay this by sacrificing guarantees of optimality) and 
($2$) an inability to learn from previous experiences, instead requiring the search to start from scratch for each instance.

\noindent\textbf{Learning-Based Methods.} 
\cite{he2014l2b, gasse2019exactoptimizationgraph, liang2024bignn} use imitation learning and graph networks to assist the pruning process of exact methods. 
\cite{park2023schedulenet, cao2022dan} solve the routing problem end-to-end by using graph networks to sequentially assign tasks to idle agents. 
These methods show effective scaling for fleet sizes between five and twenty, but they rely on a centralized solver with global state information to artificially eliminate assignment conflicts. 
While \cite{park2023schedulenet} evaluates scenarios with partial fleet observability and dynamic routing, their approach relies on a centralized global task set and spatially convenient locations for spawned targets.

Hybrid architectures that combine learning-based allocation with heuristic routing have also become popular. 
\cite{guo2024imtsp} combines an attention mechanism with a graph network to allocate variable numbers of tasks, but is restricted by a fixed fleet size. 
\cite{hu2020rlhybrid} proposes a heterogeneous graph framework similar to HALO, but makes unrealistic assumptions about global observability by letting cities choose which robot should visit them regardless of the distance between them.

More recent work \cite{wang2025solvingminmaxmultipletraveling}, \cite{rodriguezcorominas2026constructmergesolve} has continued to investigate combining reinforcement learning with heuristic techniques, but only report results on the static mTSP up to 200 tasks. 
None of these approaches simultaneously considers scenarios involving multiple-depots, partial observability, and fleet sizes above 20 even though robotic fleets frequently operate under such constraints. 
%

\section{Motivating Example}
\label{sec:motivating_example}
\begin{figure}[t] 
    \centering
    \subfloat[Static and offline routing.]{
        \includegraphics[width=0.46\linewidth]{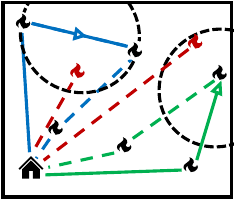}
        \label{fig:previous_motivation}
    }
    \hfill
    \subfloat[Dynamic rerouting with HALO.]{
        \includegraphics[width=0.46\linewidth]{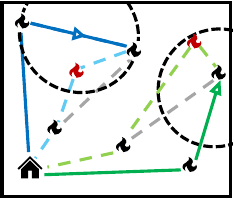}
        \label{fig:halo_motivation}
    }
    \caption{Comparison of fleet routing strategies in partially observable wildfire monitoring scenario. Black flames denote fires known a priori, while red flames indicate fires discovered during flight. (a) Traditional offline solvers require a costly second deployment (red dashed line) to visit new observations. (b) HALO enables dynamic, decentralized rerouting which dramatically improves coverage speed.}
    \label{fig:motivating_figure}
\end{figure}

Robotic fleets frequently operate in decentralized, partially observable environments due to restrictions of onboard sensors (e.g., cameras) and communication channels (e.g., radio telemetry). 
Yet, many prior works ignore such complications from the real-world. 
Thus, we illustrate the dangers of ignoring these constraints in a time-critical wildfire monitoring scenario, as many authorities leverage UAV fleets for this purpose~\cite{FAA}. 
Figure~\ref{fig:motivating_figure} shows two UAVs (blue and green triangles with observation ranges denoted by a dashed black circle) flying from a central base station to gather precise information about a spreading wildfire. 
Solid lines denote sections of the route already covered, while dashed colored lines show future paths.
Black flames represent fires known a priori to deployment while newly observed fires are shown as red flames.

By treating the routing as an offline problem, current solutions provide the fleet with a static plan that does not react to new observations. 
When a UAV observes a new fire (in red color), as in Figure \ref{fig:previous_motivation}, it can make one of two decisions: ($1$) return to the base station to relay the new observed fire so that the solver can recompute the routing solution or ($2$) ignore the observation and continue the route as initially planned. 
In this case, the UAVs continue on their planned route and require a second trip (red dashed line) to visit the newly observed fires.
This increases energy expenditure and mission makespan, and delays the rapid intervention required for wildfire scenarios. 

To tackle the limitation of the static and offline routing algorithms in Figure~\ref{fig:previous_motivation}, HALO considers decentralized and multi-depot mechanics. 
Instead of relying on a distant base station for task allocation and routing, robots using the HALO framework dynamically alter their own routes onboard to accommodate new observations (e.g., fires). 
Figure \ref{fig:halo_motivation} shows two UAV rerouting their trajectories mid flight to include new observations. 
This dramatically increases the coverage speed of tasks that are unknown a priori without requiring a centralized global update for the entire fleet.

\section{Problem Formulation}
\label{sec:problem formulation}
We consider a fleet of $M$ robots that must complete a set of $N$ tasks in a 2D environment. 
A robot $m_i$ at time $i$ is defined by the tuple $\langle d_i, l_i,\omega_i \rangle$, where $d_i$ represents the depot location, $l_i$ denotes the current location, and $\omega_i$ is $m_i$'s local observation given an observation radius $r$. 
Because the robots lack access to a global state, we define the VRP as a ND-POMDP~\cite{nair2005ndpomdp}. 
Robots communicate through a dynamic network graph $G_n(t) = (M, \mathcal{E}(t))$, where an edge exists between neighboring robots if the distance between them is less than a communication threshold $\tau$.

The fleet's primary objective is to minimize the maximum time required by any one robot to complete its route, also known as the makespan. 
We define the trajectories of the fleet as $\Sigma = [\sigma_1, \sigma_2, \dots, \sigma_M]$ given a discrete allocation $A$, where $\sigma_i$ represents the ordered list of tasks assigned to robot $m_i$. 
This Min--Max objective balances the workload across the fleet in an attempt to avoid idle agents.

The communication graph $G_n(t)$ expands into a larger heterogeneous graph $G(t)$ that is used in the learning process. 
This graph contains edges between homogeneous pairs of nodes based on the communication radius, and edges between heterogeneous pairs of nodes based on the observation radius. 
We define the system state $s_t$ as the feature representation of this heterogeneous graph at time $t$, which serves as the direct input to the policy $\pi_\theta$ described in Section \ref{sec:methodology}. 

To bridge our framework with the operational challenges discussed in Section \ref{sec:motivating_example}, and to compare our proposed method with previous works (e.g. ScheduleNet \cite{park2021schedulenet}, DAN \cite{cao2022dan}), we evaluate HALO across two distinct scenarios: 

\noindent\textbf{Single-Depot VRP.}
In this scenario, we mirror the traditional routing problem where a central base station is assumed to have perfect, instantaneous communication with all members of the fleet like in autonomous warehouse fulfillment. 
All robots start at the same location ($d_1=d_2=\dots=d_M$) and the local communication and observation radii, $\tau$ and $r$, become infinite. 
This relaxes the network-distributed and partial observability constraints resulting in a fully connected heterogeneous graph, $G$.

\noindent\textbf{Online VRP with Multiple Depots.}
This problem models the time-critical, network-distributed environments representative of the wildfire monitoring scenario in Section \ref{sec:motivating_example}. 
Robots may reside at unique depots ($d_1\ne d_2\ne\dots\ne d_M$) and the communication and observation radii, $\tau$ and $r$, become limited.
Unlike static routing problems, the task set in this online variant dynamically changes as robots complete existing tasks.  
Because we do not enforce a centralized, sequential decision-making process, robots must propagate information through the communication graph to resolve assignment conflicts and restructure routes.

\section{Methods}
\label{sec:methodology}

\begin{figure*}[!t]
    \centering
    \includegraphics[width=\textwidth]{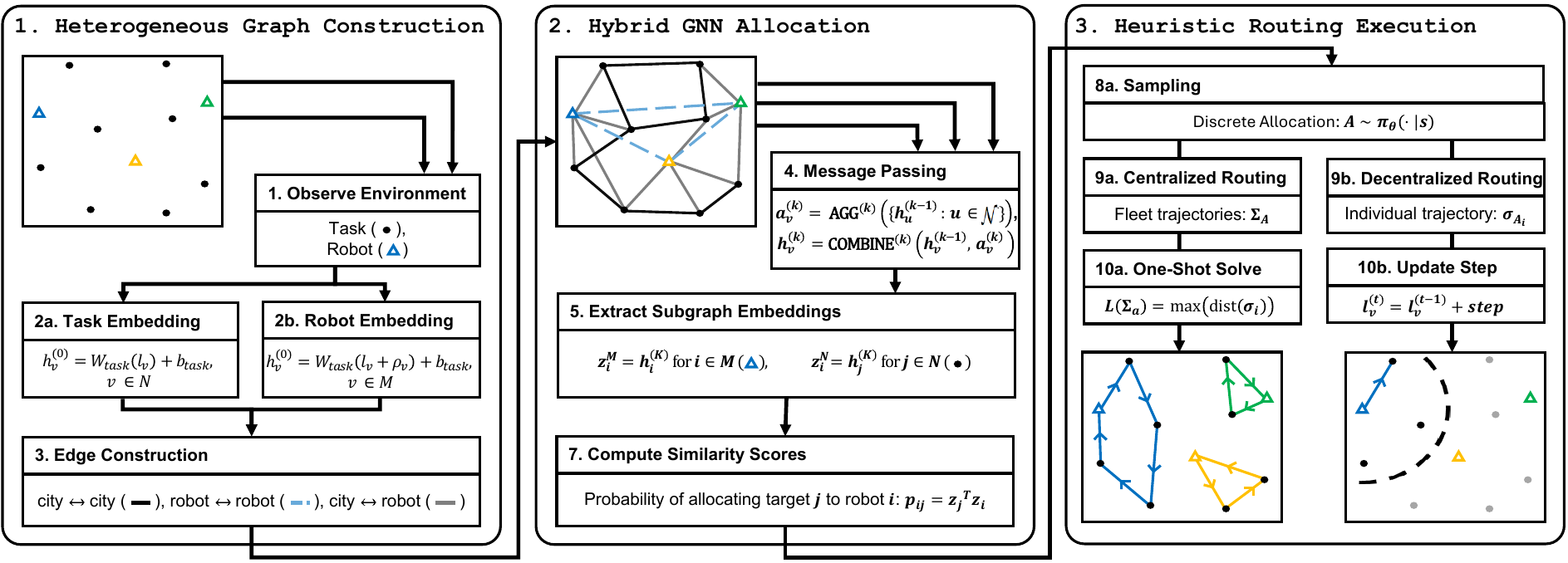} 
    \caption{Overview of HALO's work flow and architecture consisting of three different components: (\texttt{\textbf{1}}) Heterogeneous Graph Construction, (\texttt{\textbf{2}}) Hybrid GNN Allocation, and (\texttt{\textbf{3}}) Heuristic Routing Execution.} 
    \label{fig:forward_halo}
\end{figure*}

We outline how a decentralized robot using HALO  (\texttt{\textbf{1}}) builds its \textit{dynamic local subgraph} based on observation and communication ranges $r$ and $\tau$, (\texttt{\textbf{2}}) utilizes a \textit{multi-channel heterogeneous GNN} with distinct message passing layers to create a discrete task allocation, and (\texttt{\textbf{3}}) uses the heuristic solver to generate routing solutions as shown in Figure \ref{fig:forward_halo}. Finally, we discuss the reinforcement learning optimization and curriculum strategy used to train the network.

\subsection{Dynamic Local Subgraphs}
\label{subsec:subgraphs}
During decentralized execution, a robot $m_i$ develops a \textit{dynamic local subgraph} at time $t$ based on its current location $l_i$ and sensor ranges $r$ and $\tau$. 
To process this data, the robot embeds the initial node features into a high-dimensional latent space using a simple linear layer. 
We define the embedding function for a node $v$ as: 
\begin{equation} 
f_{embed}(v) =
    \begin{cases}
        W_{n}(l_v) + b_{n}, & \text{if } v \in N \\
        W_{m}(l_v + \rho_v) + b_{m}, & \text{if } v \in M
    \end{cases}
\end{equation}

\noindent where $l_v$ denotes the location of the node and $\rho$ represents a density feature that we define as the percentage of known tasks that $m_i$ is closest to. 
Because the fleet lacks a centralized controller, this feature helps the network identify robots most likely to receive imbalanced task allocations. 
To enable HALO's heterogeneous graph solution, the local set of edges is split into three channels: homogeneous spatial relations (task-to-task, robot-to-robot), and heterogeneous observations (task-to-robot). 

\subsection{Multi-Channel GNN \& Heuristic Routing}
After creating the local subgraph, node representations are updated through a single layer ($k=1$) of message passing. 
While multi-hop communication can theoretically provide robots with a better understanding of the global environment, task allocation in the VRP is locally interactive. 
Thus, the difficulties of maintaining a stable multi-hop network in a dynamic environment outweigh the benefits. 
By restricting communication to immediate neighbors, robots can efficiently coordinate tasks and avoid assignment conflicts without a centralized solver.

To capture the topology with a single-hop, we incorporate a multi-channel GNN architecture that separates the learning of spatial distribution of nodes from task-to-vehicle compatibility. 
Let $h^{(0)}$ represent the initial node embeddings. 
For a given node $v \in M$, we first process robot-to-robot communication using a Graph Isomorphism Network (GIN) layer \cite{xu2019graphisomorphismnetworks}, which captures the spatial properties of the local fleet.
Shown in Equation \ref{eq:agg_com}, $\mathcal{N}_\tau(v)$ represents the robot's neighbors within communication radius $\tau$:
\begin{equation}
\label{eq:agg_com}
    a_{v_{\tau}} = \text{AGGREGATE}_{GIN}\left(\left\{
        h_u^{(0)} : u \in \mathcal{N}_{\tau}(v)
    \right\} \right)
\end{equation}
To extract the importance of unique tasks observed by the robots, we use a Graph Attention Network (GAT)~\cite{velickovic2018graphattentionnetworks} restricted to the neighborhood of tasks $\mathcal{N}_r$: 
\begin{equation}
\label{eq:agg_obs}
    a_{v_{r}} = \text{AGGREGATE}_{GAT}\left(\left\{
        h_u^{(0)} : u \in \mathcal{N}_{r}(v)
    \right\} \right)
\end{equation}
Here, the attention mechanism helps robots learn the task relevance based on their embeddings.
Aggregations are combined with the robot's initial hidden state with a mean function to ensure that no single channel overpowers the output $h_v^{(1)}$:
\begin{equation} 
h_v^{(1)} = \text{COMBINE} \left( 
    h_v^{(0)}, a_{v_{\tau}}, a_{v_{r}}
\right)
\end{equation}

For a given node $v \in N$, message passing occurs along the spatial edges constructed in Section \ref{subsec:subgraphs} based on the observed locations of the tasks. 
Thus, we also use a GIN layer to learn neighborhood geometry:
\begin{equation}
\label{eq:agg_neigh}
    a_{v_{neigh}} = \text{AGGREGATE}_{GIN}\left(\left\{
        h_u^{(0)} : u \in \mathcal{N}_{neigh}(v)
    \right\} \right)
\end{equation}
\begin{equation}
\label{eq:task_combination}
    h_v^{(1)} = \text{COMBINE} \left(
        h_v^{(0)}, a_{v_{neigh}}
    \right)
\end{equation}
Equation \ref{eq:task_combination} also uses a mean function to combine the task's initial hidden state and neighborhood information into a final node embedding $h_v^{(1)}$.
%

After the message passing step, the node embeddings are projected to a shared latent space to compute pairwise compatibility scores between vehicles and tasks.
We compute the assignment probability matrix $P$ as the softmax over the vehicle dimension.
Each task $j$ in the local view of the ego robot has a probability $p_{ij}$ to be assigned to a robot $m_i$, where $m_i$ is the set of nearby robots including itself.

By reducing the multi-robot VRP into a set of single-agent routing problems, our hybrid framework allows us to leverage the consistency and speed of a heuristic solver~\cite{ortoolsrouting} which excels at small-scale, single-agent problems. 
Let $\mathcal{T}(\cdot)$ denote the deterministic TSP solver that maps an allocation $A$ to a set of robot trajectories, $\Sigma = \mathcal{T}(A)$, where $\Sigma$ represents the ordered routes for the local fleet. 
During inference, robots use a greedy sampling strategy to select the highest probability task allocations before computing their individual routes. 
This strategy induces coordination between neighbors as each agent predicts the most likely paths of each neighbor in its communication radius. 

\subsection{Optimization Approach}
To train the heterogeneous GNN, we utilize a reinforcement learning approach alongside a curriculum strategy to train the heterogeneous graph network described above. 
To encourage exploration of the solution space during training, we stochastically sample task allocations from the policy distribution, $\pi_\theta(A | s)$. 
Because the stochastic sampling operation is non-differentiable, we formulate the training objective as the minimization of the expected loss over the distribution of possible allocations:

\begin{equation}
J(\theta) = \mathbb{E} \big[
    {A \sim \pi_\theta} \left[ L(\Sigma) \right]
\big]
\end{equation}

\noindent Here, $L(\Sigma)$ represents the loss (makespan) of the trajectories generated from the discrete allocation:
\begin{equation}
    L(\Sigma) = \max_{i \in {1, \dots, M}} \text{dist}(\sigma_i)
\end{equation}

We compute the gradient of this objective using the Policy Gradient Theorem \cite{sutton99policygradient} which we approximate with the REINFORCE estimator \cite{williams92reinforce} such that:

\begin{equation}
\nabla_{\theta} J(\theta) \approx \frac{1}{B} \sum_{i=1}^{B} \nabla_{\theta} \log \pi_{\theta}(A_i|s_i) \left( L(\Sigma_i) - b \right)
\end{equation}

\noindent given a batch size $B$ and a baseline $b$. 
We employ a batch-average baseline, where $b = \frac{1}{B} \sum_{i=1}^B L(\Sigma_i)$. 
The baseline centers the learning signal to reduce the high variance often associated with gradient estimation. 
Allocations with shorter than average makespan produce a negative gradient term, while allocations with longer makespans have their probabilities suppressed. 

The vast state space and highly randomized configurations of a multi-depot VRP also plays a role in increasing the noisy reward signals coming from the environment.
To further stabilize the training process, we implement a curriculum learning strategy~\cite{bengio09curriculum}. 
Rather than using a complex reward structure that requires extensive tuning, the curriculum strategy guides learning by gradually expanding the spatial variance of depot locations.
\subsection{Training Details}
Training initially begins with the single-depot scenario in which depots are sampled with near-zero variance at the center of the environment. 
As training progresses, the proportion of instances that contain multi-depot scenarios linearly increases as the depot variance increases until the training reaches a fifty-fifty split.
At this point, multi-depot scenarios contain starting locations that are uniformly distributed around the environment. 
This strategy prevents catastrophic forgetting of centralized routing scenarios as the model learns to solve the multi-depot VRP. 
As a result, robots using HALO can competitively solve both static single-depot baselines and dynamic, multi-depot online VRPs without requiring two separate models.

The proposed architecture was implemented using PyTorch Geometric \cite{fey2025pyg} and was trained on a desktop workstation equipped with an Intel Core i9-14900K CPU and a single NVIDIA RTX 4080 GPU.
The hidden dimension for both task and robot embeddings was set to $d=64$. 
Network weights were updated using the Adam optimizer with a batch size of 512 and an initial learning rate of $1 \times 10^{-4}$.
During training, graph instances were dynamically generated by uniformly sampling task coordinates using a fixed random seed to ensure model generalization and strict reproducibility.
%


\begin{table*}[t]
\centering
\caption{Performance comparison on the offline, single-depot VRP at small-scale ($50$, $100$ tasks) and large-scale ($600$, $800$, $1{,}000$ tasks). $L(\Sigma)$ is the average makespan and $T(s)$ is the average time to output a solution. The symbols (--), ($\times$), and (*) denote unreported results, solution failures, and our proposed method, respectively.}
\label{tab:offline_performance}
\resizebox{\textwidth}{!}{
\begin{tabular}{l | cc | cc | cc | cc | cc}
\hline
\multicolumn{1}{c|}{} & \multicolumn{10}{c}{\# Tasks} \\
\cline{2-11}
\multicolumn{1}{c|}{Method} & \multicolumn{2}{c|}{50} & \multicolumn{2}{c|}{100} & \multicolumn{2}{c|}{600} & \multicolumn{2}{c|}{800} & \multicolumn{2}{c}{1000} \\
\cline{2-11}
 & $L(\Sigma)$ & T(s) & $L(\Sigma)$ & T(s) & $L(\Sigma)$ & T(s) & $L(\Sigma)$ & T(s) & $L(\Sigma)$ & T(s) \\
\hline
Google OR-Tools (1s) & 2.42 & 1.0 & 4.37 & 1.0 & $\times$ & $\times$ & $\times$ & $\times$ & $\times$ & $\times$ \\
Google OR-Tools (2s) & 2.12 & 2.0 & 3.79 & 2.0 & $\times$ & $\times$ & $\times$ & $\times$ & $\times$ & $\times$ \\
Google OR-Tools (1800s)~\cite{hu2020rlhybrid} & 2.03 & 3.60 & 2.27 & 36.128 & 9.64 & 1800 & 12.34 & 1800 & 14.84 & 1800 \\
\hline
ScheduleNet (g.)~\cite{park2021schedulenet} & 1.98 & -- & 2.07 & -- & -- & -- & -- & -- & -- & -- \\
ScheduleNet (s.64) & 1.92 & -- & 2.03 & -- & -- & -- & -- & -- & -- & -- \\
DAN (g.)~\cite{cao2022dan} & 2.03 & 0.30 & 2.17 & 0.48 & 3.60 & 2.58 & 4.23 & 3.36 & 4.84 & 4.21 \\
DAN (s.64) & 1.95 & 11.26 & 2.05 & 14.81 & 3.46 & 57.81 & 4.10 & 77.08 & 4.75 & 97.26 \\
\hline
Hu et al.~\cite{hu2020rlhybrid} & 1.96 & 0.02 & 2.09 & \textbf{0.04} & 3.65 & 0.81 & 4.20 & 1.69 & 4.81 & 2.87 \\
iMTSP~\cite{guo2024imtsp} & -- & -- & -- & -- & \textbf{3.42} & -- & \textbf{3.76} & -- & \textbf{4.04} & 1.98 \\
\hline
\textbf{HALO*} & \textbf{1.65} & \textbf{0.0012} & \textbf{1.92} & 0.101 & 3.62 & \textbf{1.004} & 4.08 & \textbf{1.005} & 4.57 & \textbf{1.005} \\
\hline
\end{tabular}
}
\end{table*}

\section{Evaluation}
We test each scenario against 500 VRP instances in which robots spread outwards from a central depot to complete tasks uniformly distributed in a unit square $[0,1]^2$. 
Robots in these evaluations use a single model trained on a fixed fleet size of 10 and task count of 100. 
In an effort to mirror the motivating example, we set communication radius $\tau = 0.05$ and observation radius $r = 0.3$ for all online testing. 
This simulates a fleet using an ad-hoc communication network and long-range thermal cameras to observe potential fires.

\subsection{Offline Single-Depot Results}
We benchmark HALO in the offline, single-depot variant by relaxing the constraints imposed on the robot fleet. 
In this case, robots are unrestricted by finite $\tau$ and $r$, and solve the instance in a single step. 
We evaluate against traditional heuristics~\cite{ortoolsrouting}, and four learning-based methods. 
We split these learning methods into: (1) iterative methods that build a solution sequentially~\cite{park2021schedulenet, cao2022dan}, and hybrid methods that combine a forward pass with heuristic routing~\cite{hu2020rlhybrid, guo2024imtsp}. 

For our proposed approach, total computational time includes both the models inference time and the time required to solve the single-robot subproblems. 
Performance metrics for the baseline methods are reported as published by the original authors.
\cite{park2021schedulenet} do not record results for problem sizes between 500 and 1000 nodes. 
Thus, the corresponding entries are left blank in Table \ref{tab:offline_performance}. 
We also note that \cite{hu2020rlhybrid} do not clearly define their timing methodology, thus we report their timing results with the caveat that it is uncertain whether their times account for the routing performed by a heuristic solver.
Similarly, the timing data for Guo et al. is limited, as the original text only records time for instances with $1{,}000$ tasks.

Table \ref{tab:offline_performance} shows the performance of HALO on both small-scale ($50$, $100$ tasks) and large-scale ($600$, $800$, $1{,}000$ tasks) offline problems. 
While it is clear that heuristic methods break down at large-scale, we still report the results of OR-Tools given an hour to solve from \cite{hu2020rlhybrid}.
Compared to all other tested methods, HALO records makespans $14.06\%$ shorter for problems with $50$ tasks, and $5.04\%$ shorter for problems with $100$ tasks.
Most notably, HALO constructs these improved routing results while requiring an order of magnitude less time than the heuristic methods and other learning-based methods. 

For the large-scale (i.e., $600$, $800$, and $1{,}000$ tasks) offline problems, HALO remains highly competitive against architectures optimized strictly for offline, single-depot routing scenarios while maintain the quickest solution time of all methods. 
At the largest tested problem size, iMTSP achieves an average makespan of 4.04 compared to HALO's 4.57. 
This slight trade-off in static path solutions is offset by HALO's inference speed and flexibility. 
Specialized methods like iMTSP achieve these baselines by assuming fixed fleet size and global state knowledge.
They also leave out robot locations from the embeddings, making them fundamentally incompatible with multi-depot or online VRPs. 

\begin{table*}[t]
\centering
\caption{Performance comparison on the online, multi-depot VRP across varying percentages of hidden tasks and problem sizes. $L(\Sigma)$ indicates the average makespan. Here, $T(s)$ represents the average replanning time. The symbols ($\times$), and (*) represent solution failures, and our proposed method.}
\label{tab:online_performance}
\resizebox{\textwidth}{!}{
\begin{tabular}{l | c | cc | cc | cc | cc | cc}
\hline
\multicolumn{1}{c|}{} & \multicolumn{1}{c|}{} & \multicolumn{10}{c}{\# Robots, \# Tasks} \\
\cline{3-12}
\multicolumn{1}{c|}{Method} & \multicolumn{1}{c|}{Hidden \%} & \multicolumn{2}{c|}{10, 100} & \multicolumn{2}{c|}{10, 500} & \multicolumn{2}{c|}{10, 1000} & \multicolumn{2}{c|}{50, 1000} & \multicolumn{2}{c}{100, 1000} \\
\cline{3-12}
 & & $L(\Sigma)$ & T(s) & $L(\Sigma)$ & T(s) & $L(\Sigma)$ & T(s) & $L(\Sigma)$ & T(s) & $L(\Sigma)$ & T(s) \\
\hline
OR-Tools (RH) & 0\% & 6.20 & 1.00 & 9.95 & 1.00 & 14.32 & 1.00 & $\times$ & $\times$ & $\times$ & $\times$ \\
              & 25\% & 6.30 & 1.00 & 8.12 & 1.00 & 12.25 & 1.00 & $\times$ & $\times$ & $\times$ & $\times$ \\
              & 50\% & 7.07 & 1.00 & 8.01 & 1.00 & 13.339 & 1.00 & $\times$ & $\times$ & $\times$ & $\times$ \\
\hline
\textbf{HALO (Online)*} & 0\% & 2.74 & \textbf{0.011} & 4.17 & \textbf{0.104} & 5.11 & \textbf{0.105} & \textbf{2.56} & \textbf{0.105} & \textbf{2.05} & \textbf{0.105} \\
              & 25\% & 2.71 & \textbf{0.011} & 4.23 & \textbf{0.104} & 5.14 & \textbf{0.105} & \textbf{2.59} & \textbf{0.105} & \textbf{2.08} & \textbf{0.105} \\
              & 50\% & 2.91 & \textbf{0.011} & 4.33 & \textbf{0.104} & 5.47 & \textbf{0.105} & \textbf{2.62} & \textbf{0.105} & \textbf{2.11} & \textbf{0.105} \\
\hline
HALO (Offline)* & 0\%  & \textbf{1.92} & 0.101 & \textbf{3.39} & 1.004 & \textbf{4.57} & 1.004 & 2.62 & 5.004 & 2.41 & 10.006 \\
\hline
\end{tabular}
}
\end{table*}

\subsection{Online Multi-Depot Results}
We further evaluate HALO on an online variant of the VRP in which robots are constrained by limited communication and observation radii, and tasks are incrementally spawned over time. 
We vary the percentage of tasks that begin in an unobservable state between $0\%$ and $50\%$.
Tasks only spawn as robots complete active tasks, meaning we replace each completed task with a hidden task until the hidden task list is emptied.
For example, an instance with $1{,}000$ tasks and a $50\%$ hidden percentage initializes with $500$ active tasks. 
Each time a robot completed a task, a new target dynamically spawns at a random location within the environment. 
This forces the fleet to continuously reroute based on their local observations.

This environment introduces variable fleet size and target count, multiple depots, and partial observability which baseline methods are ill-equipped to deal with. 
Methods such as those by Guo et al.~\cite{guo2024imtsp}, Park et al.~\cite{park2021schedulenet}, and Hu et al.~\cite{hu2020rlhybrid} require centralized, single-depot architectures and assume a fully observable state. 
Cao et al.'s framework~\cite{cao2022dan} relies on the global state to avoid assignment conflicts.
Thus, for these experiments, we compare our online framework against two baselines: a receding horizon implementation of the heuristic solver OR-Tools, and an offline variant of HALO without restrictions on communication and observation radii or the hidden tasks. 
This allows us to compare HALO's performance in online conditions to HALO's performance in static conditions.

The continuous rerouting in online conditions alters how computational efficiency is evaluated. 
Thus, for the online variant of HALO, we set T$(s)$ equal to the replanning time of an individual robot for a single decentralized step.
For the offline variant used in the single-depot testing, we records the total planning time as reported in Table \ref{tab:offline_performance} which sets T$(s)$ equal to the sum of time for allocation and routing.

Table \ref{tab:online_performance} demonstrates HALO's generalization across various percentages of hidden tasking, leveraging the offline variant as a comparison when HALO is unrestricted by the online environment. 
The results of the receding horizon heuristic approach demonstrate the difficulty of the task for traditional methods. 
In instances with more than 10 robots, the heuristic-based approach fails to converge, emphasizing a boundary in which traditional approaches can no longer deal with the dynamic online routing problem. 
This can also be seen in the case with $500$ tasks in which an increase in the hidden percentage actually helps the heuristic by avoiding scenarios with too many tasks. 
When operating online with 0\% hidden tasks and finite communication and observation ranges, the decentralized routes result in average makespans only 16.9\% longer than the all-knowing offline variant. 
This trade-off in performance occurs as a natural consequence of robots needing to reroute during execution as tasks enter their local observation radius. 
HALO vastly outperforms the traditional heuristic under partially observable conditions. 

HALO deals with increased variability as the environment becomes more unpredictable. 
As the proportion of hidden targets scales from $0\%$ to $50\%$, the online framework maintains highly efficient routing, reporting an average makespan increase of only $4.8\%$ across problem sizes. 
This demonstrates the architecture's ability to effectively handle dynamically spawning tasks without suffering route degradation. 
In contrast, heuristic baselines like the receding horizon OR-Tools implementation struggle under these conditions to create efficient trajectories because of their inability to learn from previous experiences and need to build routes from scratch at each iteration. 

Further, the framework scales cleanly to large robot teams, coordinating fleets of $50$ and $100$ robots without requiring any retraining of the network. 
While the overall makespan in these test cases naturally decreases due to the reduced task burden imposed on each robot, the framework's ability to process these highly variable input dimensions speaks well to its viability for massive operational areas such as the motivating wildfire example. 
We also note that by distributing the computational load throughout the execution of the routing, the online HALO framework achieves a low average replanning time of approximately $100$ ms on problems with $1{,}000$ tasks. 
This allows robots to continuously update their trajectories at high speeds which is critical for real-world applications.

\section{Conclusions and Future Work}
This paper introduces HALO, a hybrid framework with a heterogeneous graph network that separates the learning of spatial distributions and task compatibility.
Decentralized fleets of up to $100$ robots utilize their local observations to solve large vehicle routing problems in real-time.
Experimental results showed HALO outperforms state-of-the-art learning methods trained strictly for the offline, single-depot VRP on smaller scales while remaining competitive at larger problem sizes. 
HALO also thrives in more realistic, online scenarios in which robots can only partially observe the environment during deployment showing efficient routing even as the fleet must complete $1{,}000$ tasks. 
The computational and algorithmic viability of this decentralized architecture allows us to shift our focus towards physical implementation.
Future work will develop the HALO's ability to deal with increasingly complex constraints, including realistic sensor noise, dynamic physical obstacles, and adversarial interference which will help robots using HALO to conduct missions in autonomous, online scenarios. 
Ultimately, we intend to deploy and validate HALO across heterogeneous, cross-domain fleets to fully realize its potential in large-scale, real-world operations.

\addtolength{\textheight}{-3cm}   








\bibliographystyle{IEEEtran}
\bibliography{biblio}

\end{document}